\documentclass[lettersize,journal]{IEEEtran}
\usepackage{amsmath,amssymb,amsfonts}
\usepackage{algorithmic}
\usepackage{algorithm}
\usepackage{array}
\usepackage[caption=false,font=normalsize,labelfont=sf,textfont=sf]{subfig}
\usepackage{textcomp}
\usepackage{stfloats}
\usepackage{url}
\usepackage{verbatim}
\usepackage{graphicx}
\usepackage{cite}
\usepackage{CJKutf8}
\usepackage{xcolor}
\usepackage{float}
\usepackage{booktabs}   
\usepackage[colorlinks]{hyperref}  
\usepackage{multirow}   
\usepackage{stfloats}
\usepackage{makecell}
\usepackage{pifont}
\usepackage{svg} 

\begin{document}

\title{3D-SSM: A Novel 3D Selective Scan Module for Remote Sensing Change Detection}

\author{Rui Huang, Jincheng Zeng, Sen Gao, Yan Xing
\thanks{
This work was supported in part by the Fundamental Research Funds for the Central Universities, NO.3122020045. (Corresponding author:
Yan Xing.)}
\thanks{Rui Huang, Jincheng Zeng, and Sen Gao are with
the School of Computer Science and Technology, Civil Aviation University of China, Tian Jin 300300, China (e-mail: rhuang@cauc.edu.cn; jingchengzeng@163.com; aosion217@gmail.com)}
\thanks{Yan Xing is with the School of Safety Science and Engineering, Civil Aviation University of China, Tian Jin 300300, China (e-mail: yxing@cauc.edu.cn)}
}



\maketitle

\begin{abstract}
Existing Mamba-based approaches in remote sensing change detection have enhanced scanning models, yet remain limited by their inability to capture long-range dependencies between image channels effectively, which restricts their feature representation capabilities. To address this limitation, we propose a 3D selective scan module (3D-SSM) that captures global information from both the spatial plane and channel perspectives, enabling a more comprehensive understanding of the data.
Based on the 3D-SSM, we present two key components: a spatiotemporal interaction module (SIM) and a multi-branch feature extraction module (MBFEM). The SIM facilitates bi-temporal feature integration by enabling interactions between global and local features across images from different time points, thereby enhancing the detection of subtle changes. Meanwhile, the MBFEM combines features from the frequency domain, spatial domain, and 3D-SSM to provide a rich representation of contextual information within the image.
Our proposed method demonstrates favourable performance compared to state-of-the-art change detection methods on five benchmark datasets through extensive experiments. 
Code is available at \href{https://github.com/VerdantMist/3D-SSM}{https://github.com/VerdantMist/3D-SSM}.
\end{abstract}


\section{Introduction}
Remote sensing change detection (CD) aims to identify and quantify changes in coupled remote sensing images captured at the same location over a long time span that exhibit various data discrepancies, such as variations in illumination, seasonal changes, noise, and registration errors. The task of CD is challenging, as changes of interests vary significantly between different applications, ranging from tracking natural vegetation changes \cite{forest,desert,glacier} and climate change \cite{climate} to conducting disaster assessments \cite{flood,landslide,earthquake} and surveillance, as well as directing human activities \cite{crop,nc_HAN,building}.

The development of convolutional neural networks (CNNs) improved the performance of CD with various advanced modules. Daudt \emph{et al.}~\cite{daudt1} concatenate image patches from two image pairs along the channel dimension as input to a CNN and used fully connected layers to predict the probability of change within the patches. In ~\cite{daudt2}, they further propose Fully Convolutional Siamese Networks based on U-Net~\cite{unet}, enabling end-to-end CD. Within the Siamese network architecture, researchers explore dense convolution~\cite{dense1,dense2}, dilated convolution~\cite{dilated1,dilated2}, and attention mechanisms~\cite{stnet} to enhance the performance of CD algorithms. While 2D convolutions operate solely on the spatial dimensions of remote sensing images, 3D convolutional kernels offer the advantage of capturing more detailed spatio-spectral features. A representative work by Ye \emph{et al.}~\cite{3D-con-ye} propose a novel adjacent-level feature fusion network based on 3D convolutions, which leverages the inner fusion capability of 3D kernels to simultaneously extract and integrate feature information from remote sensing images. Although the convolution operation maintains a lightweight architecture with weight sharing and has excellent capability to extract shallow and local features, it struggles to capture long-term relationships between pixels and fails to capture spatially variable information within images.


To capture the global context of images, Transformers have been introduced in CD tasks, which leverage self-attention mechanisms to compute the correlations between different tokens, expanding the network's receptive field (RF). Yu \emph{et al.}~\cite{yugcformer} propose a context-aware relative position encoding method to enhance the transformer's ability to capture long-range dependencies. Song \emph{et al.}~\cite{Song2024} cascade several local–global siamese transformers to extract both local and global semantic discriminative features. Given the limited spectral information in remote sensing images, recent studies adopt channel attention and spatial attention mechanisms to separately model semantic and spatial information. For example, Cheng \emph{et al.}~\cite{cheng-isnet} introduce an improved separable network that employs channel attention to facilitate semantic-specific feature extraction and utilizes spatial attention to enhance the response to positional changes. Zhang \emph{et al.}~\cite{zhang-MSS} stack channel attention and spatial attention modules in three different configurations to better capture changes between remote sensing images. However, the inherent quadratic space and time complexity of Transformer hinders the development of more effective and computationally efficient architectures for CD.

\begin{figure*}[]
  \centering
   \includegraphics[width=0.85\linewidth]{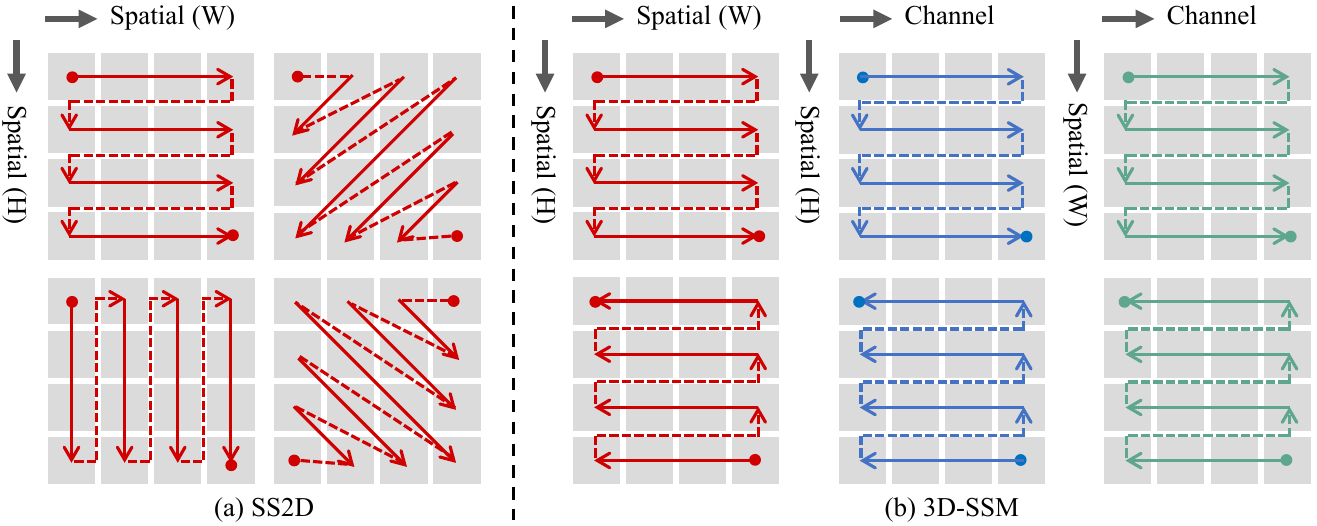}
   \caption{Illustration of SS2D and 3D-SSM.}
   \label{fig:ss3d}
\end{figure*}

Unlike Transformer, Mamba \cite{mamba} adopts state space models (SSMs) \cite{nips2021,iclr2022} and develops a selective scan mechanism (S6), which can model long-range dependencies for natural language processing (NLP) tasks with linear computational complexity. The success of Mamba in NLP accelerates the development of Mamba in computer vision area. Methods like Vision mamba \cite{zhu2024visionmamba} and Vmamba \cite{vmamba} further highlight the usability of Mamba in vision tasks. In CD area, pioneer works, like ChangeMamba \cite{changemamba} and M-CD \cite{mcd}, introduce Mamba as a feature extractor to extract global spatial contextual information from input images. 
CDMamba \cite{cdmamba} proposes a scaled residual ConvMamba (SRCM) block to extract global features and uses convolution to enhance local details. 
However, most existing Mamba-based CD methods primarily focus on spatial information modeling (Fig.~\ref{fig:ss3d} (a)), barely study the scanning model from 3D perspectives. We argue that the powerful selective scanning mechanism of Mamba can not only capture long-range dependencies between spatial pixels, but also strengthen the connections of features from different channels.

In this paper, we propose a novel scan model (Fig.~\ref{fig:ss3d} (b)), 3D selective scan module (3D-SSM), which scans image features in the HW, HC, and WC planes, where ``H", ``W", and ``C" represent height, width, and channel, respectively. 3D-SSM not only takes into account the relationships between neighboring pixels but also considers the weights across channels of each pixel. We build a spatiotemporal interaction module (SIM) and a multi-branch feature extraction module (MBFEM) as the bi-temporal feature integration and the basic feature decoder, respectively. SIM utilizes features of bi-temporal images to enhance each other and then calculates difference features for depicting changes. MBFEM combines the features of convolutional operation, Fast Fourier Transform (FFT), and 3D-SSM to extract global and local information simultaneously. We conduct extensive experiments on five remote sensing CD datasets, demonstrating the superiority of our method over the state-of-the-art CD methods.
Our main contributions are as follows:

\begin{itemize}
\item We design a 3D selective scan module (3D-SSM), for extracting more effective features for change detection by scanning features in three dimensions. 

\item We present a spatialtemporal interaction module (SIM) with 3D-SSM, which can generate effective difference features from bi-temporal image features.

\item We propose a multi-branch feature extraction module (MBFEM) to build the basic decoder, which captures contextual information to generate change estimation.

\item We have conducted various experiments on five benchmark datasets. The quantitative and qualitative results demonstrate our method outperforms other CD methods.
\end{itemize}

\section{Related Work}
\label{sec:Related Work}

\subsection{CNN-based methods}
As fundamental feature extraction methods in CD tasks, pure CNNs struggle to capture long-range dependencies and model global context. Recent efforts aim at enhancing the distinguishing ability of feature extraction by improving convolution layers \cite{nc_fullycon}, expanding the receptive field through the use of dilated convolutions \cite{cnn2} to capture global information, and emphasizing detailed edge features through hierarchical convolutions \cite{nc_HUANG}. Furthermore, Fang \emph{et al.} \cite{snunet} introduce SNUNet-CD, a densely connected Siamese network designed for extracting shallow-layer information with high-resolution and fine-grained features. Jiang \emph{et al.} \cite{cnn5} propose WRICNet which detects and fuses changes across multiple scales by adaptively weighting multi-scale features. Meshkini \emph{et al.} \cite{cnn6} employ a 3D CNN architecture to extract spatiotemporal information from continuously changing remote sensing images. Zhao \emph{et al.} \cite{sgsln} leverage dual semantic features to guide the identification and refinement of variable objects, and design a half-convolution unit to significantly reduce the computational cost of convolution operations.
Due to the inherent characteristics of CNNs, using a pure CNN as the backbone network for remote sensing image change detection tasks is evidently inefficient. In this paper, we use VMamba \cite{vmamba} as the backbone and design novel modules by jointly using convolution and Mamba to efficiently extract detailed and edge features.

\subsection{Transformer-based methods}
Transformer models have emerged as a preferred architecture for CD tasks due to their superior ability to handle long-range dependencies and learn complex patterns in images compared to CNNs. Chen \emph{et al.} \cite{bit} leverage Transformers to model contextual relationships between compact semantic token sets. Bernhard \emph{et al.} \cite{mapformer} propose MapFormer which enhances accuracy by using Transformer to incorporate pre-change semantic information. Bandara \emph{et al.} \cite{changeformer} introduce a Siamese network with a hierarchical Transformer backbone, and Zhang \emph{et al.} \cite{zhangswin} integrate a Swin Transformer within a U-shaped architecture for multi-scale feature extraction. Feng \emph{et al.} \cite{trans8} employ dual-branch processing and cross-attention for feature extraction, enhancing the model's ability to capture difference features. Jiang \emph{et al.} \cite{trans9} focus on modelling inter- and intra-relations between reliable tokens by utilizing attention schemes, emphasizing the unchanged regions. Li \emph{et al.} \cite{trans10} propose ConvTransNet, which achieves multi-scale interactive fusion between CNN and Transformer branches, improving the robustness of CD performance on the changed areas with different sizes.
Although Transformer excels at modelling long-range dependencies and parallel computation, addressing the computational complexity inherent in its attention mechanism remains a significant challenge.

\begin{figure*}[]
  \centering
  \includegraphics[width=\linewidth]{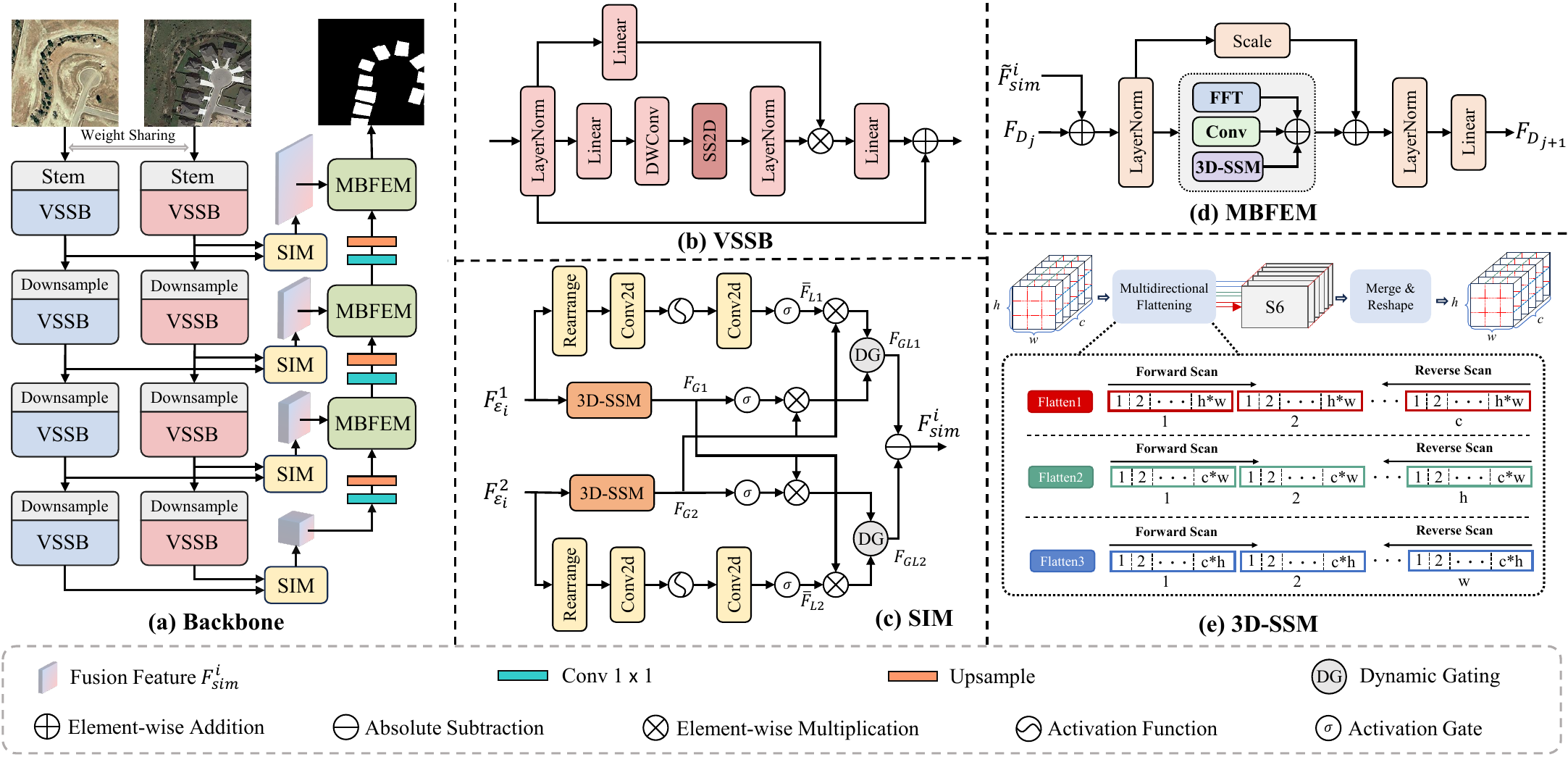}
  \caption{The main framework of the proposed method.}
  \label{fig:overview}
\end{figure*}

\subsection{Mamba-based methods}
Recently, Mamba architecture has been regarded as an effective alternative to Transformer in the visual domain due to its linear complexity, which has demonstrated its effectiveness in tasks such as image classification \cite{mambahsi}, semantic segmentation \cite{rs}, image restoration \cite{mambair,vmambair}, \emph{etc.} Mamba-based methods outperform Transformer-based models in terms of computational efficiency and accuracy, making Mamba and its variants particularly suitable for capturing contextual information from images. To capture 2D information, Chen \emph{et al.}~\cite{rsmamba} model the spatial dimensions of image features with three different scanning paths, \emph{i.e.}, forward path, reverse path, and random shuffle path, remote sensing image classification.
To capture 3D information, He \emph{et al.}~\cite{3dssmamba} propose to scan the 3D tokens along both the spatial and spectral dimensions for HSI classification.

The successful application of Mamba in the visual domain has motivated researchers to explore its potential for CD applications. Chen \emph{et al.} \cite{changemamba} are the first to explore the potential of Mamba architecture for remote sensing CD tasks, and successfully develop three tailored frameworks for binary change detection, semantic change detection, and building damage assessment, respectively. For change detection in very high-resolution (VHR) remote sensing images, Zhao \emph{et al.} \cite{rs-mamba} propose RS-Mamba, which utilizes a selective scan module to model entire large images in multiple directions directly, avoiding the loss of contextual information caused by further segmentation, and even achieving better performance on larger images. Paranjape \emph{et al.} \cite{mcd} combine multi-scale features extracted by a Mamba-based encoder in the difference module to learn temporal relations and then use the combined features for spatial learning in the decoder. To address the lack of detailed clues in CD tasks, Zhang \emph{et al.} \cite{cdmamba} introduce CDMamba, which effectively combines global and local features by leveraging the strengths of both Mamba and convolution, dynamically integrating bi-temporal features guided by another temporal image.

Unlike existing Mamba methods, we design a novel module, 3D-SSM, which handles features in 3D by scanning them in three directions. Thus, 3D-SSM is capable of capturing contextual information more comprehensively.

\section{Methodology}

\subsection{Overview}
Our purpose is to detect the changes from an image pair $\langle \textbf{I}_1,\textbf{I}_2\rangle$. As shown in Fig.~\ref{fig:overview}, the whole network adopts the Siamese network as an encoder network $\mathcal{E}$ to extract multi-scale features $\textbf{F}_{\mathcal{E}_i}^1$ and $\textbf{F}_{\mathcal{E}_i}^2$ from $\textbf{I}_1$ and $\textbf{I}_2$, respectively. We build the encoder module $\mathcal{E}_i|_{i=1}^4$ with the VSS blocks of VMamba \cite{vmamba}.
Then $\textbf{F}_{\mathcal{E}_i}^1$ and $\textbf{F}_{\mathcal{E}_i}^2$ are combined by a \textit{Spatiotemporal Interaction Module} (SIM) to generate difference feature $\textbf{F}_{\text{SIM}_i}$ to depict the changes. To efficiently compute the change map, we employ a \textit{Multi-Branch Feature Extraction Module} (MBFEM) to construct the decoder module $\mathcal{D}_j|_{j=1}^3$, which is composed of scaling, Fast Fourier Transform (FFT), convolution, and \textit{3D Selective Scan Module} (3D-SSM). MBFEM performs decoding operations on the upsampled features from the previous decoder and the SIM features at the same resolution. Thus, the decoder feature $\textbf{F}_{\mathcal{D}_j}$ not only has the ability to capture change information but also has rich low-level detail contextual information. Finally, we use a simple convolutional operation as a change classifier to produce the change map.
In the following sections, we first introduce 3D-SSM and then give the details of SIM and MBFEM, respectively. Subsequently, we present the training losses and implementation details.

\subsection{3D selective scan module (3D-SSM)}
\label{section:3d-ssm}

As shown in Fig.~\ref{fig:ss3d} (a), 2D selective scan modules, such as \cite{zhu2024visionmamba} and \cite{vmamba}, scan image features in the spatial plane to model pixel relationships across spatial dimensions. Given Mamba's linear complexity, we aim to capture more comprehensive information. Therefore, we propose a novel module, \textit{3D selective scan module} (3D-SSM), which not only performs sequence modeling along the spatial dimension of remote sensing images but also computes the channel-wise weights for each element, enabling global relationship modeling.

Given an input feature $\textbf{F}_{in} \in \mathbb{R}^{H \times W \times C}$, similar to the standard 2D selective scanning process, we first flatten it into $\textbf{X} \in \mathbb{R}^{(H\times W) \times C}$: 
\begin{equation}
    \textbf{X} = [[x_1^1,x_2^1,...,x_{h \times w}^1],...,[x_1^c,x_2^c,...,x_{h \times w}^c]],
    \label{eq:flatten}
\end{equation}
where $x^{n}_{h,w} \in \mathbb{R},\quad h \in \{1,...,H\},\ w \in \{1,...,W\},\ n \in \{1,...,C\}$ denote the feature value at height $h$, width $w$, and channel $n$.

Then we process $\textbf{X}$ and its transposed version $\textbf{X}^{\text{T}}$ with VME~\cite{zhu2024visionmamba} and combine the processed features to capture the global visual context information, which can be formulated as:
\begin{equation}
\textbf{F}_{out} = \mathcal{R}^{-1}(VME(\textbf{X}) + VME(\textbf{X}^{\text{T}})^{-\text{T}}),
\label{eq:vem}
\end{equation}
where $VME(\cdot)$ denotes VEM, $(\cdot)^{-\text{T}}$ is reverse transpose operation, $\mathcal{R}^{-1}(\cdot)$ denotes reverse reshape operation. After VME scanning, these generated 1D mapping sequences are reshaped into 3D structures awaiting subsequent fusion.

To capture the relationships in the channel dimension between adjacent pixels, as shown in Fig.~\ref{fig:overview} (e), similar to Eq.(\ref{eq:flatten}), we flatten the features into $\textbf{X}_H \in \mathbb{R}^{(W\times C) \times H}$ and $\textbf{X}_W \in \mathbb{R}^{(H\times C) \times W}$:
\begin{align}
    \textbf{X}_H = [[x_1^1,x_2^1,...,x_{w \times c}^1],...,[x_1^h,x_2^h,...,x_{w \times c}^h]]. \\
    \textbf{X}_W = [[x_1^1,x_2^1,...,x_{h \times c}^1],...,[x_1^w,x_2^w,...,x_{h \times c}^w]].
\end{align}

We then process $\textbf{X}_H$ and $\textbf{X}_W$ with Eq.~(\ref{eq:vem}) to generate $\textbf{F}_{out}^{H} $ and $\textbf{F}_{out}^{W}$, respectively. The final feature of 3D-SSM can be computed as:
\begin{equation}
    \textbf{F}_{3D} = \textbf{F}_{out} + \textbf{F}_{out}^{H} + \textbf{F}_{out}^{W}.
    \label{eq:3d-ssm}
\end{equation}

\subsection{Spatiotemporal interaction module}

Calculating the difference between the features $\textbf{F}_{\mathcal{E}_i}^1$ and $\textbf{F}_{\mathcal{E}_i}^2$ is very important for change detection. 
As shown in Fig.~\ref{fig:overview} (c), \textit{spatiotemporal interaction module} (SIM) takes bi-temporal features $\textbf{F}_{\mathcal{E}_i}^1$ and $\textbf{F}_{\mathcal{E}_i}^2$ as input, then outputs difference feature $\textbf{F}_{\text{SIM}_i}$. We use 3D-SSM to learn the global features and 2D convolution to learn the local features, which can be computed as:
\begin{align}
    &\textbf{F}_G^k = 3D\text{-}SSM(\textbf{F}_{\mathcal{E}_i}^k),\\
    &\textbf{F}_L^k = Conv_{2d}(SiLU(Conv_{2d}(\textbf{F}_{\mathcal{E}_i}^k))),
\end{align}
where $\textbf{F}_G^k$ and $\textbf{F}_L^k, k=1,2$ are global and local features extracted from $\textbf{F}_{\mathcal{E}_i}^1$ and $\textbf{F}_{\mathcal{E}_i}^2$, respectively.

Subsequently, the global feature $\textbf{F}_G^1$ is used to enhance both $\textbf{F}_G^2$ and $\textbf{F}_L^2$, while $\textbf{F}_G^2$ similarly enhances $\textbf{F}_G^1$ and $\textbf{F}_L^1$. Then we combine the enhanced features by using a dynamic gating module. The difference features $\textbf{F}_{\text{SIM}_i}$ is calculated with absolute difference. The detailed computations are as follows:
\begin{align}
    &\textbf{F}_{GL}^1 = DG(\sigma(\textbf{F}_G^2)\odot\textbf{F}_G^1,\sigma(\textbf{F}_L^2)\odot\textbf{F}_G^1),\\
    &\textbf{F}_{GL}^2 = DG(\sigma(\textbf{F}_G^1)\odot\textbf{F}_G^2,\sigma(\textbf{F}_L^1)\odot\textbf{F}_G^2),\\
    &\textbf{F}_{\text{SIM}_i} = \| \textbf{F}_{GL}^1 - \textbf{F}_{GL}^2 \|_1,
\end{align}
where $\odot$ is element-wise multiplication, $\sigma(\cdot)$ denotes softmax, $DG(\cdot,\cdot)$ represents dynamic gating operation.

\begin{figure}
    \centering
    \includegraphics[width=\linewidth]{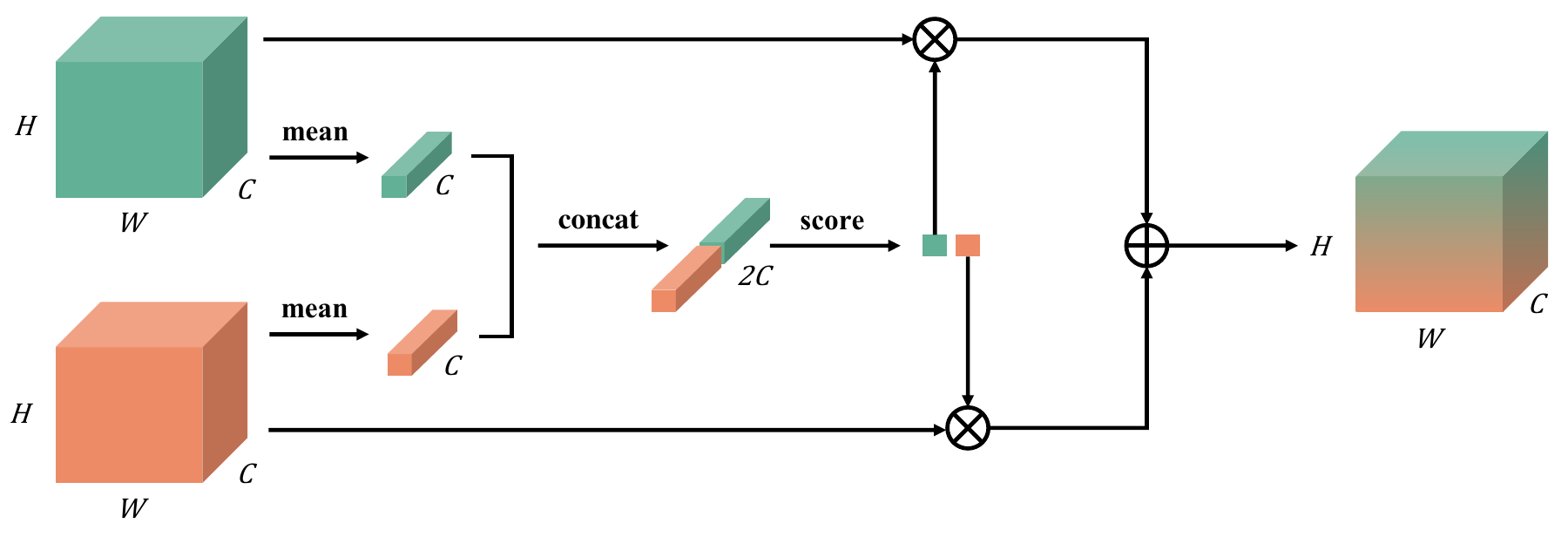}
    \caption{Illustration of the dynamic gating (DG) module in SIM.}
    \label{fig:dynamicGate}
\end{figure}

As shown in Fig.~\ref{fig:dynamicGate}, the dynamic gating operation $DG(\cdot,\cdot)$ takes two features as input. Take $\textbf{F}_1$ and $\textbf{F}_2$ as an example, the calculation process of the dynamic gating operation is as follows:
\begin{align}
    &\textbf{F} = \alpha_1 \textbf{F}_1 +\alpha_2 \textbf{F}_2,\\
    &[\alpha_1, \alpha_2] = Linear([mean(\textbf{F}_1), mean(\textbf{F}_2)]),
\end{align}
where $mean(\cdot)$ calculates the mean of each feature map along with the channel, $[\cdot, \cdot]$ denotes concatenation operation, $Linear(\cdot)$ is linear operation.

\subsection{Multi-branch feature extraction module}
As shown in Fig.~\ref{fig:overview} (d), \textit{multi-branch feature extraction module} (MBFEM) has four branches to improve feature representative ability, consisting of scaling, FFT, convolution, and 3D-SSM. We sum the upsampled feature $\textbf{F}_{D_j}$ with $\textbf{F}_{\text{SIM}_i}$ and conduct layer normalization. The normalized feature is further processed by each branch. The detailed process is as follows:
\begin{align}
\small
    &\textbf{F} = \mathcal{N}(Conv_{1\times1}(Up(\textbf{F}_{D_{j}})) + \textbf{F}_{\text{SIM}_{4-j}}),\\
    & \textbf{F}' = Linear(FFT(\textbf{F}) + Conv(\textbf{F}) + 3D\text{-}SSM(\textbf{F})),\\
    & \textbf{F}_{D_{j+1}} = Linear(\mathcal{N}(\textbf{F}' + \beta\textbf{F})),
\end{align}
where $\textbf{F}_{D_1} = \textbf{F}_{\text{SIM}_4}$ and $j = 1,2,3$. $Up(\cdot)$ denotes upsampling operation. $\beta$ is a learnable parameter. $FFT(\cdot)$ and $Conv(\cdot)$ are FFT and convolution operations. 
The FFT branch can enhance the boundaries by extracting high-frequency features. We adopt a learnable quantization matrix $\textbf{W}$ to modulate the features as:
\begin{align}
    \textbf{F}_{FFT} = \mathcal{G} ( \mathcal{P}^{-1} ( \mathcal{F}^{-1} ( \mathcal{F}(\mathcal{P}(Conv_{1 \times 1}(\mathbf{F}_{in})))\odot \mathbf{W}))),  
\end{align}
where $\mathcal{F}(\cdot)$ and $\mathcal{F}^{-1}(\cdot)$  represent FFT and inverse FFT operations, respectively. $\mathcal{P}(\cdot)$ and $\mathcal{P}^{-1}(\cdot)$ denote the patch unfolding and folding operations, respectively. $\mathcal{G}$ represents the GEGLU function from \cite{geglu}.

The convolution operation is responsible for parsing and preserving detailed information. We use two convolutional layers with $3\times3$ kernels, which can be formulated as:
\begin{equation}
    \textbf{F}_{Conv} = Conv_{3 \times 3}(SiLU (Conv_{3 \times 3}(\textbf{F}_{in}))),
\end{equation}
where $SiLU(\cdot)$ is activation function.

\subsection{Loss function}
We adopt the cross-entropy loss and Dice loss to optimize the proposed network, which are defined as:
\begin{align}
    & L_{ce}=-\frac{1}{N}\sum_{i=1}^{N}y_i\log(\hat{y}_i),\\
    & L_{dice}=1-\frac{2 {\textstyle \sum_{i=1}^{N}}y_i\hat{y}_i }{{\textstyle \sum_{i=1}^{N}}y_i+{\textstyle \sum_{i=1}^{N}}\hat{y}_i}, 
\end{align}
where $\hat{y}_i$ is predicted probability of the $i$th pixel, $y_i$ represents the corresponding groundtruth. $N$ indicates the number of pixels.

To supervise the network learning, we use simple classifiers with $1\times1$ convolutional kernels on features $\textbf{F}_{D_1}$, $\textbf{F}_{D_2}$, and $\textbf{F}_{D_3}$ to generate change masks. The total loss of our method is defined as:
\begin{equation}
    L_{total} = \sum_{j=1}^{3}(\lambda_1L_{ce}^j+\lambda_2L_{dice}^j),
    \label{eq:loss}
\end{equation}
where $\lambda_1$ and $\lambda_2$ denote the coefficients of cross entropy loss and Dice loss. We set $\lambda_1=\lambda_2 = 0.5$ in our experiment. Note that we use bi-linear upsampling to resize the change masks into the size of the groundtruth mask. The mask generated by $\textbf{F}_{D_3}$ is used as the final prediction for evaluation.

\subsection{Implementation details}
Our model is implemented by PyTorch on an NVIDIA V100 GPU. To optimize the model, we use Adam~\cite{adam2015} as the optimizer to train our model with decay rates $\beta_1 = 0.9 $ and $\beta_2 =0.999 $. The initial learning rate is set to 1e-4. We set the batch size to 12 and train our network with 100 epochs. The best model is selected for final evaluation.

Our model adopts Vmamba-S~\cite{vmamba} as the backbone. The number of VSS blocks in the four stages of the encoder is set to \{2, 2, 15, 2\}. The extracted image feature maps have spatial resolutions of 1/4, 1/8, 1/16, and 1/32 of the original image size, respectively. The number of channels is set to \{96, 192, 384, 768\}. At each stage, downsampling is performed using bilinear interpolation. In the decoding phase, the number of decoder blocks at each stage is set to \{1, 1, 1\}, with the number of channels matching those of the extracted features. Upsampling is performed using bilinear interpolation.

\begin{table*}[!ht]
  \centering
  \renewcommand{\arraystretch}{1.2}
  \caption{Quantitative results of different methods on the five CD Datasets. The best values are highlighted in \textbf{bolded}.}
  \label{tab:comparison}
  \resizebox{\textwidth}{!}{ 
  \begin{tabular}{@{}ccccccccccccccccccccccccccc@{}}
  \toprule
  \multirow{2}{*}{{Type}} & \multirow{2}{*}{{Method}} & \multicolumn{5}{c}{{WHU-CD}} & \multicolumn{5}{c}{{LEVIR-CD}} & \multicolumn{5}{c}{{CDD}} & \multicolumn{5}{c}{{SYSU-CD}} & \multicolumn{5}{c}{{DSIFN-CD}} \\
  & & Pre. & Rec. & F1 & Kappa & \multicolumn{1}{c|}{OA} & Pre. & Rec. & F1 & Kappa & \multicolumn{1}{c|}{OA} & Pre. & Rec. & F1 & Kappa & \multicolumn{1}{c|}{OA} & Pre. & Rec. & F1 & Kappa & \multicolumn{1}{c|}{OA} & Pre. & Rec. & F1 & Kappa & OA \\
  \midrule[0.7pt] 
  \multirow{5}{*}{\rotatebox{90}{\normalsize{CNN}}} & SNUNet$_{21}$ & 78.95 & 87.62 & 83.06 & 82.32 & \multicolumn{1}{c|}{98.58} & 91.98 & 90.15 & 91.06 & 90.58 & \multicolumn{1}{c|}{99.10} & 96.26 & 95.48 & 95.87 & 95.29 & \multicolumn{1}{c|}{98.98} & 82.95 & 77.50 & 80.13 & 74.27 & \multicolumn{1}{c|}{90.94} & 49.12 & 54.48 & 51.66 & 50.68 & 75.82 \\
                  & SGSLN$_{23}$ & 93.65 & 93.35 & 93.50 & 93.24 & \multicolumn{1}{c|}{99.49} & 92.37 & 90.92 & 91.64 & 91.19 & \multicolumn{1}{c|}{99.15} & 97.16 & 96.88 & 97.02 & 96.60 & \multicolumn{1}{c|}{99.27} & 84.04 & 80.47 & 82.21 & 76.88 & \multicolumn{1}{c|}{91.79} & 68.59 & 65.17 & 66.84 & 60.26 & 89.01 \\
                  & STNet$_{23}$ & 86.89 & 91.59 & 89.17 & 88.71 & \multicolumn{1}{c|}{99.12} & 92.23 & 88.94 & 90.56 & 90.06 & \multicolumn{1}{c|}{99.05} & 95.79 & 93.45 & 94.60 & 93.86 & \multicolumn{1}{c|}{98.68} & 87.43 & 76.04 & 81.34 & 76.09 & \multicolumn{1}{c|}{91.77} & 56.36 & 62.99 & 59.49 & 58.19 & 80.80 \\
                  & Changer$_{23}$(resnet50) & 92.37 & 87.48 & 89.86 & 89.45 & \multicolumn{1}{c|}{99.22} & 92.69 & 89.41 & 91.02 & 90.55 & \multicolumn{1}{c|}{99.10} & 92.03 & 86.49 & 89.18 & 87.70 & \multicolumn{1}{c|}{97.41} & 85.15 & 77.26 & 81.01 & 75.52 & \multicolumn{1}{c|}{91.46} & 65.37 & 58.85 & 61.94 & 54.64 & 87.71 \\
                  & Changer$_{23}$(resnet101) & 93.25 & 87.05 & 90.05 & 89.65 & \multicolumn{1}{c|}{99.24} & 93.10 & 88.77 & 90.88 & 90.40 & \multicolumn{1}{c|}{99.09} & 91.91 & 82.26 & 86.82 & 85.08 & \multicolumn{1}{c|}{96.92} & 86.18 & 76.76 & 81.20 & 75.83 & \multicolumn{1}{c|}{91.62} & 72.61 & 53.40 & 61.54 & 55.07 & 88.66 \\
  \midrule
  \multirow{5}{*}{\rotatebox{90}{\normalsize{Transformer}}} & BIT$_{22}$ & 89.08 & 91.74 & 90.39 & 89.99 & \multicolumn{1}{c|}{99.23} & 90.73 & 89.92 & 90.33 & 89.81 & \multicolumn{1}{c|}{99.02} & 91.77 & 91.70 & 91.74 & 90.58 & \multicolumn{1}{c|}{97.96} & 79.75 & 73.50 & 76.50 & 69.62 & \multicolumn{1}{c|}{89.35} & 66.23 & 55.70 & 60.51 & 53.25 & 87.65 \\
                  & ChangeFormer$_{22}$ & 92.52 & 83.43 & 87.74 & 87.26 & \multicolumn{1}{c|}{99.07} & 91.18 & 85.83 & 88.42 & 87.82 & \multicolumn{1}{c|}{98.86} & 92.46 & 90.05 & 91.24 & 90.02 & \multicolumn{1}{c|}{97.86} & 83.56 & 72.55 & 77.67 & 71.40 & \multicolumn{1}{c|}{90.16} & 56.28 & 70.77 & 62.70 & 53.99 & 85.69 \\
                  & ACABFNet$_{23}$ & 88.48 & 87.97 & 88.22 & 87.73 & \multicolumn{1}{c|}{99.07} & 90.20 & 87.01 & 88.57 & 87.97 & \multicolumn{1}{c|}{98.86} & 95.75 & 94.79 & 95.27 & 94.60 & \multicolumn{1}{c|}{98.84} & 82.33 & 75.65 & 78.85 & 72.68 & \multicolumn{1}{c|}{90.43} & 72.30 & 55.91 & 63.06 & 56.63 & 88.87 \\
                  & ChangeViT$_{24}$ & 95.28 & 93.50 & 94.38 & 94.15 & \multicolumn{1}{c|}{99.56} & 92.33 & \textbf{91.10} & 91.71 & 91.27 & \multicolumn{1}{c|}{99.16} & 97.02 & 96.27 & 96.64 & 96.17 & \multicolumn{1}{c|}{99.17} & 83.92 & 79.92 & 81.87 & 76.45 & \multicolumn{1}{c|}{91.65} & \textbf{77.47} & 58.52 & 66.68 & 60.97 & \textbf{90.06} \\
                  & ELGCNet$_{24}$ & 89.35 & 84.69 & 86.96 & 86.44 & \multicolumn{1}{c|}{98.99} & 90.95 & 87.33 & 89.11 & 88.53 & \multicolumn{1}{c|}{98.91} & 93.86 & 90.20 & 91.99 & 90.89 & \multicolumn{1}{c|}{98.06} & 82.20 & 75.39 & 78.65 & 72.43 & \multicolumn{1}{c|}{90.35} & 54.25 & 68.66 & 60.61 & 51.38 & 84.84 \\
  \midrule
  \multirow{5}{*}{\rotatebox{90}{\normalsize{Mamba}}} & ChangeMamba$_{24}$ & \textbf{96.22} & 93.53 & 94.85 & 94.65 & \multicolumn{1}{c|}{99.60} & 93.01 & 90.46 & 91.71 & 91.28 & \multicolumn{1}{c|}{99.16} & 95.98 & 96.41 & 96.19 & 95.65 & \multicolumn{1}{c|}{99.06} & 85.65 & 81.79 & 83.68 & 78.79 & \multicolumn{1}{c|}{91.67} & 67.78 & 72.31 & 69.97 & 63.59 & 89.20 \\
                  & RS-Mamba$_{24}$ & 91.81 & 86.94 & 89.31 & 88.88 & \multicolumn{1}{c|}{99.17} & 92.17 & 90.28 & 91.21 & 90.75 & \multicolumn{1}{c|}{99.11} & 95.92 & 96.33 & 96.12 & 95.58 & \multicolumn{1}{c|}{99.04} & 79.17 & 76.68 & 77.91 & 71.23 & \multicolumn{1}{c|}{89.74} & 55.15 & 44.76 & 49.42 & 40.33 & 84.43 \\
                  & CDMamba$_{24}$ & 94.35 & 91.85 & 93.08 & 92.80 & \multicolumn{1}{c|}{99.46} & 91.91 & 90.41 & 91.15 & 90.15 & \multicolumn{1}{c|}{99.06} & 95.82 & 94.50 & 95.15 & 94.48 & \multicolumn{1}{c|}{98.81} & 81.19 & 77.06 & 79.07 & 72.83 & \multicolumn{1}{c|}{90.38} & 59.45 & 59.47 & 59.46 & 51.16 & 86.22 \\
                  & M-CD$_{24}$ & 95.69 & 94.17 & 94.92 & 94.71 & \multicolumn{1}{c|}{99.60} & 92.99 & 90.66 & 91.81 & 91.38 & \multicolumn{1}{c|}{99.18} & \textbf{97.79} & 97.33 & 97.56 & 97.22 & \multicolumn{1}{c|}{99.40} & \textbf{88.80} & 77.94 & 83.01 & 78.21 & \multicolumn{1}{c|}{92.48} & 64.17 & 69.34 & 66.66 & 59.51 & 88.21 \\
                  & Ours & 96.06 & \textbf{94.58} & \textbf{95.31} & \textbf{95.12} & \multicolumn{1}{c|}{\textbf{99.63}} & \textbf{93.36} & 90.92 & \textbf{92.12} & \textbf{91.71} & \multicolumn{1}{c|}{\textbf{99.21}} & 97.72 & \textbf{97.81} & \textbf{97.77} & \textbf{97.45} & \multicolumn{1}{c|}{\textbf{99.45}} & 86.17 & \textbf{83.36} & \textbf{84.90} & \textbf{80.32} & \multicolumn{1}{c|}{\textbf{92.98}} & 67.95 & \textbf{73.03} & \textbf{70.40} & \textbf{64.07} & 89.56 \\
  \bottomrule
  \end{tabular}
  }
\end{table*}

\begin{figure*}[!ht]
  \centering
  \includegraphics[width=\linewidth]{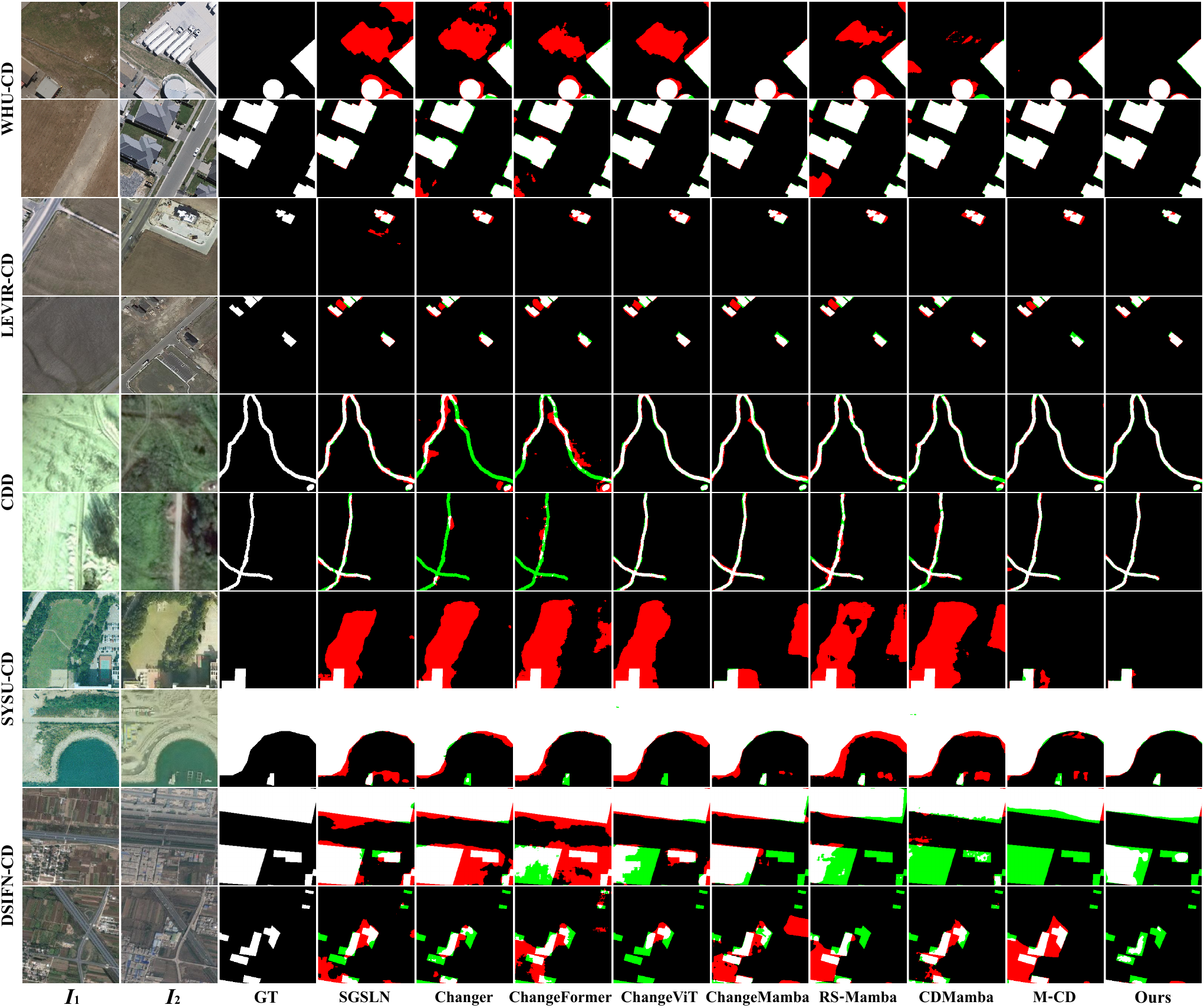}
  \caption{Typical change detection results on five datasets. White represents true positives, black represents true negatives, green represents false positives and red represents false negatives.}
  \label{fig:visualization}
\end{figure*}


\section{Experiment}
\subsection{Setup}

\subsubsection{Baselines}
We select thirteen state-of-the-art change detection (CD) methods as baselines, comprising four CNN-based approaches: SNUNet \cite{snunet}, SGSLN \cite{sgsln}, STNet \cite{stnet}, and Changer \cite{changer}; five Transformer-based CD methods: BIT \cite{bit}, ChangeFormer \cite{changeformer}, ACABFNet \cite{acabfnet}, ChangeViT \cite{changevit}, and ELGCNet \cite{noman2024elgcnet}; and four Mamba-based CD methods: Changemamba \cite{changemamba}, RS-Mamba \cite{rsmamba}, CDMamba \cite{cdmamba}, and M-CD \cite{mcd}. 
All compared methods were retrained 100 epochs using the official implementation and default hyperparameters, ensuring a fair comparison on the benchmark datasets.

\subsubsection{Datasets}
We adopt five widely used remote sensing change detection datasets for comparison, including LEVIR-CD \cite{LEVIR}, WHU-CD \cite{WHU}, CDD \cite{CDD}, SYSU-CD \cite{SYSU} and DSIFN-CD \cite{DSIFN}. The details of the datasets are as follows.

LEVIR-CD \cite{LEVIR} comprises high-resolution (0.5 meters per pixel) Google Earth image pairs, capturing bi-temporal changes over 9 years (spanning from 5 to 14 years). The dataset encompasses diverse building types and significant land-use transformations, featuring 10,192 image pairs in total: 7,120 for training, 1,024 for validation, and 2,048 for testing.

WHU-CD \cite{WHU} dataset is a curated subset of the larger WHU Building Dataset, designed specifically for building change detection tasks. Comprising paired aerial images from 2012 and 2016, this dataset captures changes in a region affected by an earthquake, where reconstruction efforts and new construction have taken place. In line with the official protocol, the dataset is partitioned into training (5,947 pairs), validation (743 pairs), and test sets (744 pairs).

CDD \cite{CDD} comprises high-resolution remote sensing image pairs captured by Google Earth (DigitalGlobe) that exhibit seasonal changes over the same geographical area. The dataset is particularly well-suited for detecting alterations to large-scale infrastructure, such as buildings, vehicles, and changes in natural vegetation patterns over time. Each pair consists of 256$\times$256 crop segments extracted from the original images, resulting in a total of 16,000 image pairs across three sets: 10,000 in the training set, 3,000 in the validation set, and 3,000 in the test set.

The SYSU-CD dataset \cite{SYSU}, comprising 20,000 pairs of high-resolution aerial images (0.5 meters per pixel) captured between 2007 and 2014 over Hong Kong, provides an ideal platform for evaluating change detection (CD) algorithms in challenging urban and coastal environments. This dataset is particularly well-suited for assessing CD methods due to the presence of complex shadows and bias effects resulting from high-rise building development and infrastructure growth. Specifically, SYSU-CD includes 12,000 image pairs for training, 4,000 for validation, and a separate set of 4,000 test pairs to ensure unbiased evaluation.

The DSIFN-CD dataset \cite{DSIFN} was curated from Google Earth, comprising six large bi-temporal high-resolution image sets covering six cities in China. Each original pair of images, measuring 512$\times$512 pixels, was divided into four equal-sized sub-pairs (256$\times$256), ensuring consistency with prior datasets. The resulting dataset contained a substantial training set with 14,400 pairs, as well as validation and test sets comprising 1,360 and 192 pairs, respectively.

\subsubsection{Criteria}
We adopt Precision (Pre.), Recall (Rec.), F1 score (F1), Kappa and Overall Accuracy (OA) to evaluate different change detectors. Precision reflects the proportion of true positive pixels among all the pixels predicted as positive. Recall indicates the proportion of true positive pixels in the ground truth out of all actual positive pixels. The F1 score balances precision and recall by calculating their harmonic mean. The Kappa coefficient measures the agreement between the predicted changes and ground truth while considering the possibility of random matches, providing a more balanced assessment than simple accuracy. OA represents the proportion of correctly predicted pixels out of the total pixels.

\subsection{Result analysis}
Table~\ref{tab:comparison} presents quantitative results for various change detectors on five benchmark datasets. The results indicate that Mamba-based methods generally outperform those based on Transformers and CNNs, underscoring the ability of Mamba to extract higher-quality features than its counterparts. Within the CNN-based approaches, SGSLN achieves the highest F1 scores across all five datasets. Meanwhile, among Transformer-based methods, ChangeViT yields the best performance. Notably, M-CD stands out as the top-performing Mamba-based method, outshining others in this category. Except for ChangeMamba and M-CD, SGSLN outperforms other CNN- and Transformer-based methods on CDD, SYSU-CD, and DSIFN-CD, while ChangeViT excels on WHU-CD and LEVIR-CD. Compared to SGSLN, our method demonstrates significant F1 score improvements across all datasets: 1.81\% on WHU-CD, 0.48\% on LEVIR-CD, 0.75\% on CDD, 2.69\% on SYSU-CD, and 3.56\% on DSIFN-CD. Additionally, our method outperforms M-CD with notable F1 score increments: 0.39\% on WHU-CD, 0.31\% on LEVIR-CD, 0.21\% on CDD, 1.89\% on SYSU-CD, and 3.74\% on DSIFN-CD. The experimental results show that our approach yields superior performance compared to alternative methods in multiple change detection benchmarks.

Fig.~\ref{fig:visualization} presents typical change detection results on various datasets. On WHU-CD, CNN- and Transformer-based methods (e.g., SGSLN and ChangeViT) miss-classify unwanted changes into changes. While Mamba-based methods achieve less false negative detections than CNN- and Transformer-based methods.
On LEVIR-CD, all compared methods struggle with small change object and complex boundary, except for our approach. 
On CDD, Changer and ChangeFormer perform poorly in detecting changes along narrow paths, and all compared methods introduce significant boundary errors.
Large color variations in the SYSU-CD lead to false detections in background regions for most methods except our method. On DSIFN-CD, our method reduces false negatives in dense urban building scenes, outperforming other methods. In summary, our method achieves more accurate boundary detection under various scenarios and demonstrates stronger robustness to color variations compared to other methods.


\begin{table}[]
  \centering
  \caption{Comparison of the number of parameters, FLOPs, and inference time of an image pair with size of $256\times256$.}
  \label{tab:param&count}
    \begin{tabular}{c|c|ccc}
    \toprule
    Type        & Method        & Params.(M)  & FLOPs(G) & Time(ms) \\ \midrule
    \multirow{4}{*}{\rotatebox{90}{CNN}}         & SNUNet      & 3.01     & 55.14    & 46.98 \\
                & SGSLN          & 6.04        & 46.02    & 18.95 \\
                & STNet          & 14.62       & 38.45    & 29.03 \\
                & Changer        & 11.35       & 23.10    & 48.57 \\ \midrule
    \multirow{5}{*}{\rotatebox{90}{Transformer}} & BIT         & 3.5      & 42.53    & 20.77 \\
                & ChangeFormer   & 41.03       & 202.86   & 29.84 \\
                & ACABFNet       & 102.32      & 113.14   & 40.68 \\
                & ChangeViT      & 32.14       & 155.24   & 32.14 \\
                & ELGCNet        & 10.57       & 751.94   & 32.28 \\ \midrule
    \multirow{5}{*}{\rotatebox{90}{Mamba}}       & ChangeMamba & 49.94    & 114.82   & 98.30 \\
                & RS-Mamba       & 42.30       & 73.33    & 30.03 \\
                & CDMamba        & 8.88        & 216.74   & 108.63 \\
                & M-CD           & 51.23       & 91.26    & 114.93 \\
                & Ours           & 10.63       & 562.33   & 104.92 \\
    \bottomrule
    \end{tabular}
\end{table}

Table~\ref{tab:param&count} shows the number of parameters, FLOPs, and the average inference time. Our method has the smallest parameter number among all compared Mamba-based change detection methods, excluding CDMamba. Notably, despite having the second largest FLOPs, we achieve faster inference times than both CDMamba and M-CD.

\begin{figure*}[!ht]
  \centering
  \includegraphics[width=\linewidth]{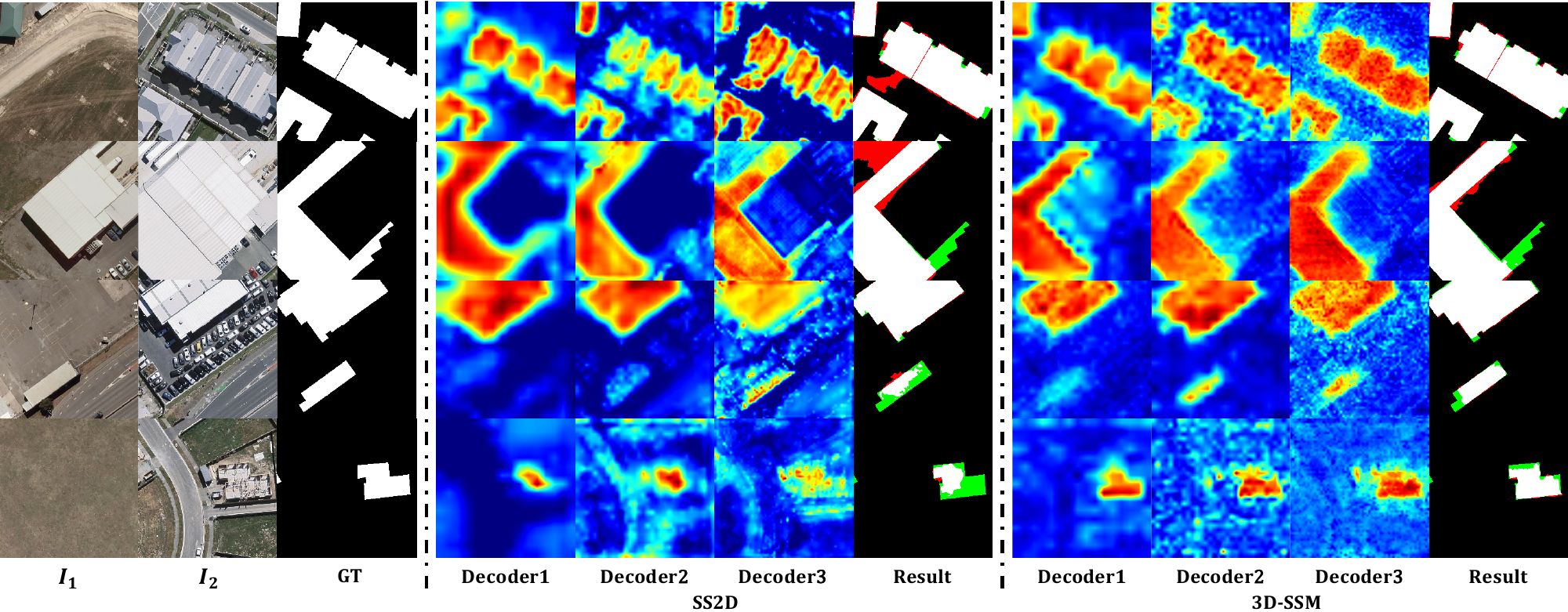}
  \caption{Feature visualization of MBFEM using SS2D or 3D-SSM. The boundaries of 3D-SSM are more accurate than those of SS2D.}
  \label{fig:feature_vis}
\end{figure*}

\begin{figure*}[!ht]
  \centering
  \includegraphics[width=\linewidth]{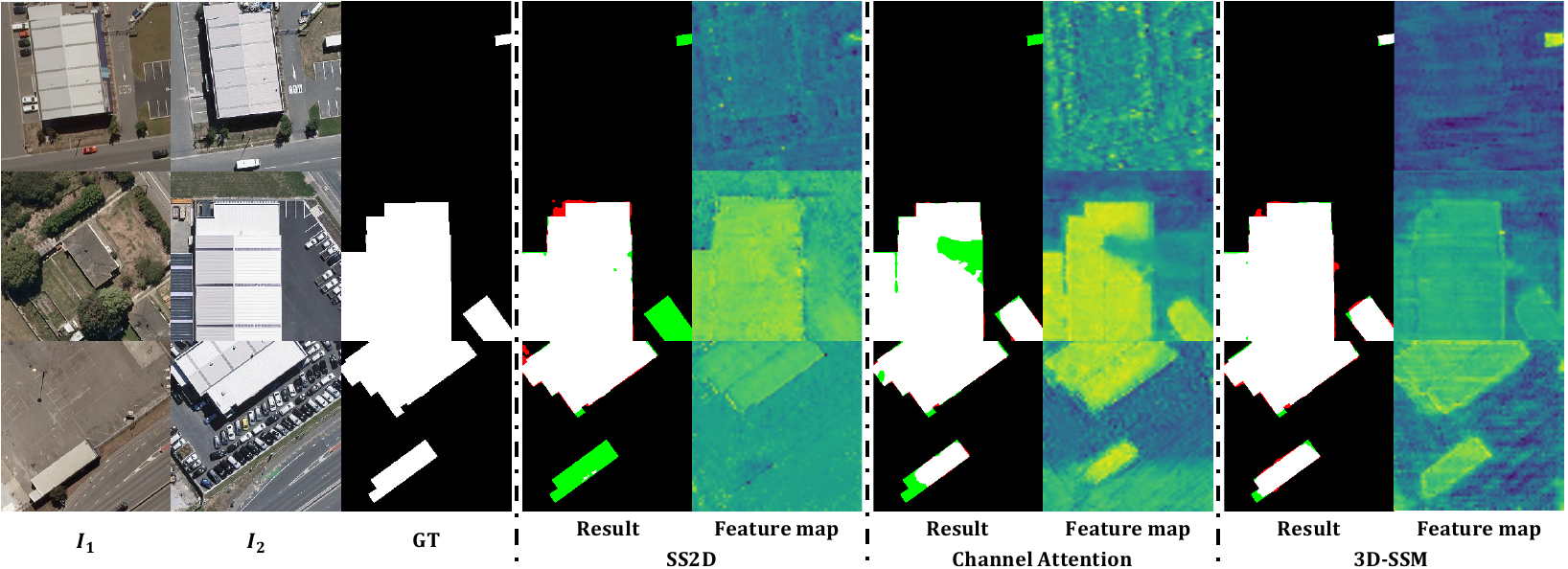}
  \caption{Visualization of the predicted results and the feature maps of replacing 3D-SSM with SS2D and channel attention of MBFEM. The feature maps are from the fourth branch at the Decoder3. 3D-SSM can capture minute changes (see the first pair of images), provide more complete regions (see the second pair of images), and achieve more precise boundaries (see the third pair of images).}
  \label{fig:feature_vis_branch}
\end{figure*}

\subsection{Ablation studies}

\subsubsection{Effectiveness of 3D-SSM}
To evaluate the effectiveness of 3D-SSM, we conduct ablation studies by replacing it with SS2D and SS2D + channel attention. As shown in Table \ref{tab:ablation_ss3d}, the results indicate that adding channel attention to SS2D 
improves the F1 score by 0.1\% and 0.27\% on WHU-CD and SYSU-CD datasets, respectively. In contrast, using 3D-SSM boosts the F1 score to 95.3\% and 84.9\%, and increases OA to 99.63\% and 92.98\% on WHU-CD and SYSU-CD datasets, respectively. The experimental results demonstrate that 3D-SSM is effective for CD. Visualizing the feature maps of the MBFEM in Figs.~\ref{fig:feature_vis} and \ref{fig:feature_vis_branch} further supports this finding. The boundaries of the feature maps generate using 3D-SSM are more accurate than those obtained with SS2D, resulting in detected change regions that are closer to the ground truth.

\begin{table}
    \centering
    \caption{Comparison of the performance of our method using SS2D or 3D-SSM. ``CA'' denotes channel attention. The best values are highlighted in \textbf{bolded}.}
    \resizebox{\linewidth}{!}{
      \begin{tabular}{c|cccc|cccc}
        \toprule
        \multirow{2}{*}{Methods} & \multicolumn{4}{c}{WHU-CD} & \multicolumn{4}{c}{SYSU-CD} \\           
        & Pre. & Rec. & F1 & OA & Pre. & Rec. & F1 & OA \\ \midrule
        SS2D    & \textbf{97.09} & 93.10          & 95.05          & 99.62          & \textbf{87.59} & 81.74 & 84.56 & 92.96 \\
        SS2D+CA & 95.88          & \textbf{94.44} & 95.15          & 99.62          & 86.06 & \textbf{83.63} & 84.83 & 92.95 \\
        3D-SSM  & 96.56          & 94.07          & \textbf{95.30} & \textbf{99.63} & 86.17 & 83.36 & \textbf{84.90} & \textbf{92.98} \\ 
        \bottomrule
      \end{tabular}
    }
    \label{tab:ablation_ss3d}
\end{table}

\subsubsection{Scanning model in 3D-SSM}
We conduct an ablation study in Table~\ref{tab:ablation of view} to study the impact of the selective scanning models of Mamba from different perspectives on the final CD results. When only using the HW perspective (i.e., SS2D), we achieve the highest precision values on WHU-CD and SYSU-CD. However, the corresponding lower recall values affect their F1 scores.
Without HW, scanning from HC or WC perspectives yields the lowest F1 scores on both datasets.
An intriguing observation is that combining the HW perspective with either HC or WC does not yield any significant improvement and resulted in reduced F1 scores.
In contrast, scanning from both HC and WC perspectives together demonstrate mutual complementarity between row and column data, improving F1 scores by 0.17\% and 0.06\% relative to SS2D on WHU-CD and SYSU-CD, respectively. Integrating all three perspectives achieved the most comprehensive feature representation, achieving the highest F1 score of 95.30\% and 84.90\% on WHU-CD and SYSU-CD, respectively.

\begin{table}[!t]
    \centering
    \caption{Impact of scanning perspectives on change detection performance. The best values are highlighted in \textbf{bolded}.}
    \resizebox{\linewidth}{!}{
        \begin{tabular}{@{}ccc|cccc|cccc@{}}
        \toprule
        \multicolumn{3}{c}{Perspective} & \multicolumn{4}{c}{WHU-CD} & \multicolumn{4}{c}{SYSU-CD}\\
        HW & HC & WC & Pre. & Rec. & F1 & OA & Pre. & Rec. & F1 & OA \\
        \midrule
        \checkmark &  &  & \textbf{97.09} & 93.10 & 95.05 & 99.62 & \textbf{88.00} & 81.80 & 84.79 & \textbf{93.08} \\
         & \checkmark &  & 95.57 & 94.04 & 94.80 & 99.59 & 84.45 & 84.43 & 84.44 & 92.66 \\
         &  & \checkmark & 96.14 & 93.79 & 94.95 & 99.60 & 83.37 & \textbf{85.95} & 84.64 & 92.64 \\
        \checkmark & \checkmark &  & 95.89 & 94.14 & 95.01 & 99.61 & 87.40 & 81.55 & 84.38 & 92.88 \\
         & \checkmark & \checkmark & 96.62 & 93.87 & 95.22 & \textbf{99.63} & 86.06 & 83.68 & 84.85 & 92.95 \\
        \checkmark &  & \checkmark & 95.80 & \textbf{94.31} & 95.05 & 99.61 & 87.87 & 81.75 & 84.70 & 93.03 \\
        \checkmark & \checkmark & \checkmark & 96.56 & 94.07 & \textbf{95.30} & \textbf{99.63} & 86.17 & 83.36 & \textbf{84.90} & 92.98 \\
        \bottomrule
        \end{tabular}
    }
    \label{tab:ablation of view}
\end{table}

\subsubsection{Effects of different components in MBFEM}
Table~\ref{tab:ablation of modules} reports the results of our method using MBFEM with different components.
Only using FFT, our method achieves the lowest F1 scores of 95.14\% and 84.20\% on WHU-CD and SYSU-CD, respectively.
On WHU-CD, combining FFT with 3D-SSM results in a 0.12\% decrease in the F1 score compared to using 3D-SSM alone, while on SYSU-CD, combining FFT with convolution leads to a 0.11\% decrease compared to using convolution alone.
In other cases, using two different components simultaneously in MBFEM can improve the performances of CD. Notably, when the three components (Conv, FFT, and 3D-SSM) are used together in MBFEM, the highest F1 scores are achieved on WHU-CD and SYSU-CD datasets, demonstrating their synergistic benefits.

\begin{table}
    \centering
    \caption{Ablation study of different components of MBFEM. The best values are highlighted in \textbf{bolded}.}
    \resizebox{\linewidth}{!}{
        \begin{tabular}{@{}ccc|cccc|cccc@{}}
        \toprule
        \multicolumn{3}{c}{Modules} & \multicolumn{4}{c}{WHU-CD} & \multicolumn{4}{c}{SYSU-CD} \\
         Conv & FFT & 3D-SSM & Pre. & Rec. & F1 & OA & Pre. & Rec. & F1 & OA \\
        \midrule
          \checkmark &  &  & 96.19 & 94.14 & 95.16 & 99.62 & 87.13 & 82.18 & 84.58 & 92.93 \\
          & \checkmark &  & \textbf{96.92} & 93.42 & 95.14 & 99.53 & \textbf{89.07} & 79.84 & 84.20 & 92.93 \\
          &  & \checkmark & 94.91 & \textbf{95.52} & 95.21 & 99.62 & 86.95 & 82.39 & 84.61 & 92.93 \\
          \checkmark & \checkmark &  & 96.34 & 94.11 & 95.21 & 99.62 & 84.87 & 84.08 & 84.47 & 92.71 \\
          \checkmark & & \checkmark & 96.69 & 93.89 & 95.27 & 99.56 & 87.04 & 82.79 & 84.86 & \textbf{93.04} \\
          & \checkmark & \checkmark & 96.56 & 93.66 & 95.09 & 99.52 & 82.86 & \textbf{86.78} & 84.65 & 92.58 \\
          \checkmark & \checkmark & \checkmark & 96.56 & 94.07 & \textbf{95.30} & \textbf{99.63} & 86.17 & 83.36 & \textbf{84.90} & 92.98 \\
        \bottomrule
        \end{tabular}
    }
    \label{tab:ablation of modules}
\end{table}

\subsubsection{Effectiveness of SIM}
To verify the effectiveness of SIM, we replace SIM with simple absolute difference operation, i.e., $|\mathbf{F}_{\mathcal{E}_i}^1-\mathbf{F}_{\mathcal{E}_i}^2|$. As shown in Table~\ref{tab:comparison of SIM}, using SIM results in higher F1 and OA values on both datasets compared to using absolute difference. Specifically, using SIM achieves 0.37 and 0.47 F1 improvements, and 0.03 and 0.15 OA improvements over using absolute difference, respectively. 
These results demonstrate that our proposed SIM can achieve better difference features by interacting global and local representations of bi-temporal features than naive operation, e.g., absolute difference.

\begin{table}
    \centering
    \caption{Comparison of the SIM with Absolute Difference Operation. ``AD" denotes Absolute Difference. The best values are highlighted in \textbf{bolded}.}
    \resizebox{\linewidth}{!}{
      \begin{tabular}{c|cccc|cccc}
        \toprule
        \multirow{2}{*}{Method} & \multicolumn{4}{c}{WHU-CD} & \multicolumn{4}{c}{SYSU-CD} \\           
        & Pre. & Rec. & F1 & OA & Pre. & Rec. & F1 & OA \\ \midrule
        AD  & 95.52 & \textbf{94.34} & 94.93 & 99.60 & \textbf{86.57} & 82.39 & 84.43 & 92.83 \\
        SIM & \textbf{96.56} & 94.07 & \textbf{95.30} & \textbf{99.63} & 86.17 & \textbf{83.36} & \textbf{84.90} & \textbf{92.98} \\
        \bottomrule
      \end{tabular}
    }
    \label{tab:comparison of SIM}
\end{table}

\subsubsection{Effects of using different loss weight}
We introduce a hybrid loss function in Eq.(\ref{eq:loss}), where \(\lambda_1\) and \(\lambda_2\) are used to adjust the relative contributions of the BCE loss and the Dice loss to the overall loss. Table~\ref{tab:ablation_lossweight} shows the results of our method trained by Eq.(\ref{eq:loss}) with different weights. On both WHU-CD and SYSU-CD, the combination of \(\lambda_1 = 0.5\) and \(\lambda_2 = 0.5\) achieve the highest F1 scores, outperforming all other setups. Thus, we set \(\lambda_1 = \lambda_2 = 0.5\) in all our experiments.

\begin{table}
    \centering
    \caption{Ablation study of different weight for loss. The best values are highlighted in \textbf{bolded}.}
    \resizebox{\linewidth}{!}{
      \begin{tabular}{@{}cc|cccc|cccc@{}}
        \toprule
        \multicolumn{2}{c}{Weights} & \multicolumn{4}{c}{WHU-CD} & \multicolumn{4}{c}{SYSU-CD} \\
         $\lambda_1$ & $\lambda_2$ & Pre. & Rec. & F1 & OA & Pre. & Rec. & F1 & OA \\ \midrule
        0    & 1    & 93.93 & 93.84          & 93.89          & 99.52          & 86.85 & 82.22 & 84.47 & 92.87 \\
        0.25 & 0.75 & 95.88 & \textbf{94.13} & 94.99          & 99.61          & 84.71 & \textbf{84.02} & 84.36 & 92.65 \\
        0.5  & 0.5  & \textbf{96.56} & 94.07 & \textbf{95.30} & \textbf{99.63} & 86.17 & 83.36 & \textbf{84.90} & 92.98 \\ 
        0.75 & 0.25 & 96.50 & 94.03          & 95.25          & 99.63          & \textbf{88.50} & 81.49 & 84.85 & \textbf{93.14} \\
        1    & 0    & 95.62 & 94.28          & 94.94          & 99.60          & 87.68 & 80.98 & 84.20 & 92.83 \\
        \bottomrule
      \end{tabular}
    }
    \label{tab:ablation_lossweight}
\end{table}

\subsubsection{Effects of using different backbones}
In Table~\ref{tab:ablation_backbone}, we compare the variants of our model by using different backbones on WHU-CD. Notably, VMamba-S achieves SOTA results while having the smallest model parameters among all options. Furthermore, leveraging pretrained models consistently obtains a substantial improvements on F1 scores. It suggests that \ding{182} Mamba-based backbone is capable of extracting more representative features than CNN- or Transformer-based backbones; \ding{183} Using pretrained model can improve the CD performance.

To further prove the effectiveness of 3D-SSM, we replace SS2D with 3D-SSM in VSSB of Vmamba-S~\cite{vmamba}. Note that due to the limitation of the GPU memory, we set the batch size to 8. We train the corresponding models on WHU-CD without pretraining.
The F1 score of 3D-SSM based model achieves 0.74\% F1 improvement over VMamba-S based model, which demonstrates that 3D-SSM is superior than SS2D as the basic feature extraction module.

\begin{table}[]
\centering
    \caption{Ablation study on different backbones.}
    \resizebox{\linewidth}{!}{
        \begin{tabular}{cccccc}
        \toprule
        Backbone & Pretrained & Batch-size & Param(M) & GFLOPs & F1 \\
        \midrule
        \multirow{2}{*}{Resnet-50} & w/o & \multirow{2}{*}{12} & \multirow{2}{*}{32.55} & \multirow{2}{*}{723.89} & 86.57 \\
                                            & w/ & & & & 92.97 \\
        \multirow{2}{*}{ViT-S}     & w/o & \multirow{2}{*}{12} & \multirow{2}{*}{53.75} & \multirow{2}{*}{1031.04} & 90.58 \\
                                            & w/ & & & & 93.51 \\
        \multirow{2}{*}{VMamba-S}    & w/o & \multirow{2}{*}{12} & \multirow{2}{*}{10.63} & \multirow{2}{*}{562.33} & 90.91 \\
                                            & w/ & & & & 95.30 \\
        \midrule
        \multirow{2}{*}{VMamba-S} & w/o & \multirow{2}{*}{8} & \multirow{2}{*}{10.63} & \multirow{2}{*}{281.17} & 90.09 \\
                                            & w/ & & & & - \\
        \multirow{2}{*}{3D-SSM}     & w/o & \multirow{2}{*}{8} & \multirow{2}{*}{73.34} & \multirow{2}{*}{503.21} & 90.83 \\
                                            & w/ & & & & - \\
        \bottomrule
        \end{tabular}
    }
    \label{tab:ablation_backbone}
\end{table}

\subsection{Limitation}

While our method achieves promising results in five datasets, there are several limitations which should be discussed. \ding{182} Long inference time. As shown in Table~\ref{tab:param&count}, we can find that Mamba-based methods take longer inference times than CNN- or Transformer-based methods. \ding{183} Large FLOPs. The main building block, 3D-SSM scans features from three perspectives, resulting in larger FLOPs than most of the compared methods. \ding{184} As demonstrated in Figs.~\ref{fig:feature_vis} and \ref{fig:feature_vis_branch}, although our features are better than the features of SS2D and Channel Attention, we still get wrong detections at the boundaries of the change objects.

To address these challenges, future works will focus on optimizing the model to strike a better balance between performance and computational efficiency. This may involve exploring techniques such as model pruning and quantization. Additionally, parallel processing techniques or more efficient scanning algorithms can be used to maintain the high performance of 3D-SSM while significantly reducing computational overhead and inference time.

\section{Conclusion}
In this paper, we introduce a novel feature extraction module called \textit{3D Selective Scan Module} (3D-SSM). This innovative module enables simultaneous scanning of image features in three planes. As a result, 3D-SSM can capture complex relationships between spatial and channel dependencies. Leveraging the strengths of 3D-SSM, we develop two key components: the \textit{Spatiotemporal Interaction Module} (SIM) and the \textit{Multi-Branch Feature Extraction Module} (MBFEM). SIM effectively fuses temporal image features and computes difference features, which are then progressively fused by MBFEM decoders to generate a change mask. Through experiments on five benchmark datasets with 13 state-of-the-art change detectors and ablation studies on the WHU-CD dataset, we demonstrate that our proposed method outperforms existing approaches.

\bibliographystyle{IEEEtran} 
\bibliography{refs}{}

\end{document}